\documentclass[10pt,twocolumn,letterpaper]{article}

\usepackage{iccv}
\usepackage{times}
\usepackage{epsfig}
\usepackage{graphicx}
\usepackage{amsmath}
\usepackage{multirow}
\usepackage{amssymb}
\usepackage[ruled,vlined]{algorithm2e}
\usepackage{caption}
\usepackage{subcaption}
\usepackage{amsmath}
\usepackage{adjustbox}
\usepackage{multirow}
\usepackage{nomencl}
\usepackage{mathtools}  
\usepackage{amsmath}
\usepackage{amssymb}
\usepackage{tabulary}
\usepackage{booktabs}
\usepackage{amsmath,lipsum,xcolor,caption}
\makenomenclature
\usepackage{verbatim}
\usepackage{subcaption}
\usepackage[pagebackref=true,breaklinks=true,letterpaper=true,colorlinks,bookmarks=false]{hyperref}

\iccvfinalcopy % *** Uncomment this line for the final submission

\def\iccvPaperID{27} % *** Enter the ICCV Paper ID here

\ificcvfinal\fi

\begin{document}

%%%%%%%%% TITLE
\title{AdvGAN++ : Harnessing latent layers for adversary generation}

\author{Puneet Mangla\thanks{ Authors contributed equally}\\
IIT Hyderabad, India\\
{\tt\small cs17btech11029@iith.ac.in}
\and
Surgan Jandial$^{*}$ \\
IIT Hyderabad, India\\
{\tt\small jandialsurgan@gmail.com}
\and
Sakshi Varshney$^{*}$ \\
IIT Hyderabad, India\\
{\tt\small cs16resch01002@iith.ac.in}
\and
Vineeth N Balasubramanian\\
IIT Hyderabad, India\\
{\tt\small vineethnb@iith.ac.in}
}

\maketitle
% Remove page # from the first page of camera-ready.
\ificcvfinal\thispagestyle{empty}\fi

%%%%%%%%% ABSTRACT
\begin{abstract}
Adversarial examples are fabricated examples, indistinguishable from the original image that mislead neural networks and drastically lower their performance. Recently proposed AdvGAN, a GAN based approach, takes input image as a prior for generating adversaries to target a model. In this work, we show how latent features can serve as better priors than input images for adversary generation by proposing AdvGAN++, a version of AdvGAN that achieves higher attack rates than AdvGAN and at the same time generates perceptually realistic images on MNIST and CIFAR-10 datasets.
\end{abstract}

%%%%%%%%% BODY TEXT
\section{Introduction and Related Work}
Deep Neural Networks(DNNs), now have become a common ingredient to solve various tasks dealing with classification, object recognition, segmentation, reinforcement learning, speech recognition etc. However recent works \cite{xie2017adversarial,10.1007/978-3-319-66399-9_4,DBLP:journals/corr/abs-1805-07820,Sharif:2016:ACR:2976749.2978392,DBLP:journals/corr/abs-1807-03326,DBLP:journals/corr/HuangPGDA17} have shown that these DNNs can be easily fooled using carefully fabricated examples that are indistinguishable to original input. Such fabricated examples, knows as adversarial examples mislead the neural networks by drastically changing their latent features, thus affecting their output. \par
Adversarial attacks are broadly classified into \textit{\textbf{White box}} and \textit{\textbf{Black box}} attacks. \textit{\textbf{White box}} attacks such as FGSM \cite{Goodfellow} and DeepFool \cite{Seyed-Mohsen} have access to the full target model. In contrary to this  \textit{black box} attacks like  Carlini and Wagner. \cite{Carlini}, the attacker does not have access to the structure or parameters of the target model, it only has access to the labels assigned for the selected input image.
\par Gradient based attack methods like Fast Gradient Sign Method (FGSM) obtains an optimal max-norm constrained perturbation of 
\begin{equation}
\eta = \epsilon  sign ( \triangledown _x J ( \theta,x,y ) )
\end{equation}
where J is the cost function and gradient is calculated w.r.t to input example.\par 
Optimization-based methods like Carlini Wagner \cite{Carlini} optimize the adversarial perturbations subject to several constraints. This approach targets  $L_{0}$, $L_{2}$, $L_{\infty}$ distance metrics for attack purpose. The optimization objective used in the approach makes it slow as it can focus on one perturbation instance at a time.
\par
In contrary to this, AdvGAN \cite{advgan} used a GAN \cite{GAN} with an encoder-decoder based generator to generate perceptually more realistic adversarial examples, close to original distribution. The generator network produces adversarial perturbation $G(x)$ when an original image instance $(x)$ is provided as input. The discriminator tries to distinguish adversarial image $(x+G(x))$ with original instance $(x)$. Apart from standard GAN loss, it uses hinge loss to bound the magnitude of maximum perturbation and an adversarial loss to guide the generation of image in adversarial way. Though, AdvGAN is able to generate the realistic examples, it fails to exploit latent features as priors which are shown to be more susceptible to the adversarial perturbations recently \cite{Abhishek}.
\par
Our Contributions in this work are:
\begin{itemize}
\item{ We show that the latent features serve as a better prior for adversarial generation than the whole input image for the untargeted attacks thereby utilizing the observation from \cite{Abhishek} and at same time eliminating the need to follow encoder-decoder based architecture for generator, thus reducing training/inference overhead.
}
\item{Since GANs are already found to work well in a conditioned setting \cite{pix2pix,cgan}, we show that we can directly make generator to learn the transition from latent feature space to adversarial image rather than from the whole input image.}
\end{itemize}
 
In the end, through quantitative and qualitative evaluation we show that our examples look perceptually very similar to the real ones and have higher attack success rates compared to AdvGAN.

\section{Methodology}
\subsection{Problem definition}
Given a model $M$ that accurately maps image $x$ sampled from a distribution $p_{data}$ to its corresponding label $t$, We train a generator $G$ to generate an adversary $x_{adv}$ of image $x$ using its feature map (extracted from a feature extractor) as prior. Mathematically :
\begin{equation}
    x_{adv} = G(z|f(x)) 
\end{equation} such that
\begin{equation}
     M(x_{adv}) \neq t ,
\end{equation}
\begin{equation}
     \Vert x - x_{adv} \Vert _{p} < \epsilon,
\end{equation}
where $1 \leq p < \infty , \epsilon > 0 $, $f$ represents a feature extractor and $\epsilon$ is maximum magnitude $\Vert.\Vert_{p}$ perturbation allowed.
\subsection{Harnessing latent features for adversary generation}
\begin{figure}
    \centering
    \includegraphics[scale=0.45]{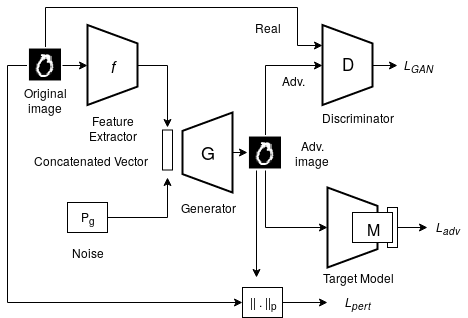}
    \caption{AdvGAN++ architecture.}
    \label{fig:architecture}
\end{figure}

We now propose our attack, AdvGAN++ which take latent feature map of original image as prior for adversary generation. Figure \ref{fig:architecture} shows the architecture of our proposed network. It contains the target model $M$ , a a feature extractor $f$, generator network $G$ and a discriminator network $D$. The generator $G$ receives feature $f(x)$ of image $x$ and a noise vector $z$ (as a concatenated vector) and generates an adversary $x_{adv}$  corresponding to $x$. The discriminator $D$ distinguishes the distribution of generator output with actual distribution $p_{data}$. In order to fool the target model $M$, generator minimize $M_t(x_{adv})$, which represents the softmax-probability of adversary $x_{adv}$ belonging to class $t$. To bound the magnitude of perturbation, we also minimize $l_2$ loss between the adversary $x_{adv}$ and $x$. The final loss function is expressed as :
\begin{equation}
    L(G,D) = L_{GAN} + \alpha L_{adv} + \beta L_{pert}
\end{equation}
where
\begin{equation}
    L_{GAN} = E\textsubscript{x}[logD(x) +E\textsubscript{x}log(1- D(G(z|f(x)))],
\end{equation}
\begin{equation}
    L_{adv} = E\textsubscript{x} [M_t(G(z|f(x)))],
\end{equation}
\begin{equation}
    L_{pert} = E\textsubscript{x}\Vert x - G(z|f(x)) \Vert _{2}
\end{equation}

Here $\alpha$ , $\beta$ are hyper-parameters to control the weight-age of each objective. The feature $f(.)$ is extracted from one of the intermediate convolutional layers of target model $M$. By solving the min-max game $arg\min_G\max_D L(G,D)$ we obtain optimal parameters for $G$ and $D$. The training procedure thus ensures that we learn to generate adversarial images close to input distribution that harness the susceptibility of latent features to adversarial perturbations. Algorithm \ref{algorithm} summarizes the training procedure of AdvGAN++.\par
\begin{algorithm}[htb]
  \SetAlgoLined
     \For {number of training iterations}{
      Sample a mini-batch of $m$ noise samples \{ $z^{(1)}$, ... $z^{(m)}$ \} from noise prior $p_g(z)$ \; \texttt{\\}
      Sample a mini-batch of $m$ examples \{$x^{(1)}$, ... $x^{(m)}$ \} from data generating distribution $p_{data}(x)$\; \texttt{\\}
      Extract latent features \{$f(x^{(1)})$, ... $f(x^{(m)})$ \}\; \texttt{\\}

     Update the discriminator by ascending its stochastic gradient. \;
     $\triangledown _{\theta _D} \frac{1}{m} \sum_{i=1}^{m} log (D ( x^{(i)} )) + log ( 1-D( G ( z^{(i)}| f(x^{(i)}))))$\; \texttt{\\}
     Sample a mini-batch of $m$ noise samples \{ $z^{(1)}$ ,$z^{(2)}$ ... $z^{(m)}$ \} from noise prior $p_g(z)$\; \texttt{\\}
     Update the generator by descending its stochastic gradient. \;
    $\triangledown _{\theta _G} \frac{1}{m} \sum_{i=1}^{m} log(1- D (G ( z^{(i)}| f(x^{(i)})))$ \ $+ \Vert x^{(i)} -G ( z^{(i)}| f(x^{(i)})) \Vert _2 + M_t(G ( z^{(i)}| f(x^{(i)})))  $
     }
     \caption{AdvGAN++ training}
     \label{algorithm}

\end{algorithm}
\begin{table*}
\begin{center}
\scalebox{0.9}{
\hspace{-1cm}
\begin{tabular}{|c|c|c|c|c|}
\hline 
\textbf{Data} & \textbf{Model} & \textbf{Defense} & \textbf{AdvGAN} & \textbf{AdvGAN++} \\
\hline\hline
\multirow{3}{*}{MNIST} & \multirow{3}{*}{Lenet C} & FGSM Adv. training &  18.7 & \textbf{20.02}\\
& & Iter. FGSM training &  13.5 & \textbf{27.31}\\
& & Ensemble training &  12.6 & \textbf{28.01}\\
\hline
\multirow{6}{*}{CIFAR-10} & \multirow{3}{*}{Resnet-32} & FGSM Adv. training & 16.03 & \textbf{29.36}\\
& & Iter. FGSM training & 14.32 & \textbf{32.34} \\
& & Ensemble training  & 29.47 & \textbf{34.74}\\ \cline{2-5}
& \multirow{3}{*}{Wide-Resnet-34-10} & FGSM Adv. training &  14.26 & \textbf{26.12}\\
& & Iter. FGSM training &  13.94& \textbf{43.2} \\
& & Ensemble training &  20.75 & \textbf{23.54}\\
\hline
\end{tabular}}
\end{center}
\caption{Attack success rate of Adversarial examples generated AdvGAN++ when target model is under defense. }
\label{table:under-defense}
\end{table*}

\section{Experiments}
In this section we evaluate the performance of AdvGAN++, both quantitatively and qualitatively. We start by describing datasets and model-architectures followed by implementation details and results.\par
\textbf{Datasets and Model Architectures}: We perform experiments on MNIST\cite{mnist} and CIFAR-10\cite{cifar} datasets wherein we train AdvGAN++ using training set and do evaluations on test set. We follow Lenet architecture C from \cite{tramer2017ensemble} for MNIST\cite{mnist} as our target model. For CIFAR-10\cite{cifar}, we show our results on Resnet-32 \cite{resnet} and Wide-Resnet-34-10 \cite{wide-resnet}. \par 
% We first evaluate the attack success rate o various defense
% mechanism such as FGSM[5] adversarial training (Adv.),
% iterative FGSM training (Iter. Adv.)[12], ensemble adver-
% sarial training (Ens.)[19]. We show that our approach per-
% forms better than the existing attack mechanism under the
% defense environment. 
\begin{figure*}[!htbp]
\centering
  \begin{subfigure}[b]{\textwidth}
\minipage{0.25\textwidth}
  \includegraphics[width=4.5cm,height=4.5cm]{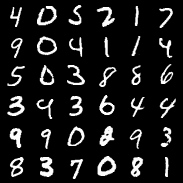}
  %\caption{UI of mobile application - side menu and main menu buttons}\label{fig:ui1}
\endminipage\hspace{6mm}
\minipage{0.25\textwidth}
  \includegraphics[width=12cm,height=4.5cm]{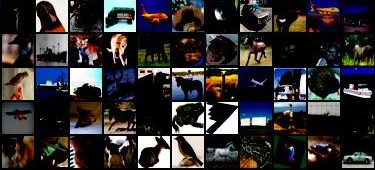}
  %\caption{Prototype of local controller along with DIGW and ampere sensors}\label{fig:local_controller}
\endminipage\hfill

  %\caption{DIGW used in real-time deployment of SPACE}\label{fig:a}
\end{subfigure}
% \end{figure*}

% \begin{figure*}[!htbp]
 \begin{subfigure}[b]{\textwidth}

\minipage{0.25\textwidth}
  \includegraphics[width=4.5cm,height=4.5cm]{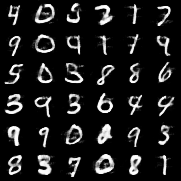}
  %\caption{UI of mobile application - side menu and main menu buttons}\label{fig:ui1}
\endminipage\hspace{6mm}
\minipage{0.25\textwidth}
  \includegraphics[width=12cm,height=4.5cm]{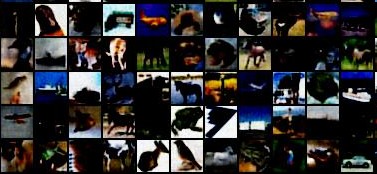}
  %\caption{Prototype of local controller along with DIGW and ampere sensors}\label{fig:local_controller}
\endminipage\hfill

  %\caption{DIGW used in real-time deployment of SPACE}\label{fig:b}
\end{subfigure}
  \caption{ Adversarial images generated by AdvGAN++ for MNIST and CIFAR-10 dataset. Row 1: Original image, Row 2: generated adversarial example. }
  \label{fig:visual-results}
\end{figure*}
\subsection{Implementation details}
We use an encoder and decoder based architecture of discriminator $D$ and generator $G$ respectively. For feature extractor $f$ we use the last convolutional layer of our target model $M$. Adam optimizer with learning rate 0.01 and $\beta_1$ = 0.5 and $\beta_2$ = 0.99 is used for optimizing generator and discriminator. We sample the noise vector from a normal distribution and use label smoothing to stabilize the training procedure.
\subsection{Results}
\textbf{Attack under no defense} We compare the attack success rate of examples generated by AdvGAN and AdvGAN++ on target models without using any defense strategies on them. The results in table \ref{table:no-defense} shows that with much less training/inference overhead, AdvGAN++ performs better than AdvGAN.\par

\begin{table}
\begin{center}
\scalebox{0.9}{
\begin{tabular}{|c|c|c|c|c|c|}
\hline
\textbf{Data} & \textbf{Target Model} & \textbf{AdvGAN} & \textbf{AdvGAN++} \\
\hline\hline
MNIST & Lenet C & 97.9 & \textbf{98.4}  \\ \hline
\multirow{2}{*}{CIFAR-10} 
& Resnet-32 & 94.7 & \textbf{97.2} \\
& Wide-Resnet-34-10 & 99.3 & \textbf{99.92} \\
\hline
\end{tabular}}
\end{center}
\caption{Attack success rate of AdvGAN and AdvGAN++ under no defense}
\label{table:no-defense}
\end{table}

\textbf{Attack under defense} We perform experiment to compare the attack success rate of AdvGAN++ with AdvGAN when target model $M$ is trained using various defense mechanism such as FGSM\cite{Goodfellow} , iterative FGSM \cite{DBLP:journals/corr/KurakinGB16} and ensemble adversarial training \cite{tramer2017ensemble}. For this, we first generate adversarial examples using original model $M$ as target (without any defense) and then evaluate the attack success rate of these adversarial examples on same model, now trained using one of the aforementioned defense strategies. Table \ref{table:under-defense} shows that AdvGAN++ performs better than the AdvGAN under various defense environment.\par

\textbf{Visual results} Figure \ref{fig:visual-results} shows the adversarial images generated by AdvGAN++ on MNIST\cite{mnist} and CIFAR-10\cite{cifar} datasets. It shows the ability of AdvGAN++ to generate perceptually realistic adversarial images.\par

\textbf{Transferability to other models} Table \ref{table:transfer} shows attack success rate of adversarial examples generated by AdvGAN++ and evaluated on different model $M^{'}$ doing the same task. From the table we can see that the adversaries produced by AdvGAN++ are significantly transferable to other models performing the same task which can also be used to attack a model in a black-box fashion. \par

\begin{table}
\begin{center}
\scalebox{0.8}{
\begin{tabular}{|c|c|c|c|c|c||c|c|}
\hline 
\textbf{Data} & \textbf{Target Model} & \textbf{Other Model} & \textbf{Attack Success rate} \\
\hline\hline
MNIST & LeNet C & LeNet B \cite{tramer2017ensemble} & 20.24 \\
\hline
\multirow{2}{*}{CIFAR-10} & Resnet-32 A & Wide-Resnet-34 & 48.22 \\
& Wide-Resnet-34  & Resnet-32 & 89.4 \\
\hline
\end{tabular}}
\end{center}
\caption{Transferability of adversarial examples generated by AdvGAN++}
\label{table:transfer}
\end{table}

\section{Conclusion}
In our work, we study the gaps left by AdvGAN \cite{advgan} mainly focusing on the observation \cite{Abhishek} that latent features are more prone to alteration by adversarial noise as compared to the input image. This not only reduces training time but also increases attack success rate.
This vulnerability of latent features made them a better candidate for being the starting point for generation and allowed us to propose a generator that could directly convert latent features to the adversarial image.

{\small
\bibliographystyle{ieee}
\bibliography{egbib}
}

\end{document}